\documentclass[10pt,twocolumn,letterpaper]{article}
\usepackage{wacv}
\usepackage{times}
\usepackage{epsfig}
\usepackage{graphicx}
\usepackage{amsmath}
\usepackage{amssymb}
\usepackage{booktabs}
\usepackage{multirow}
\usepackage{bbm}
\usepackage{epigraph}
\usepackage{multicol}
\usepackage{caption}
\usepackage{mathtools}
\usepackage[normalem]{ulem}
\usepackage{hyperref}
\hypersetup{
    colorlinks=true,
    urlcolor=magenta,
}

\usepackage[accsupp]{axessibility} 
\newcommand{\model}{VL-MPAG Net}

\def\wacvPaperID{413} 

\wacvalgorithmstrack   

\wacvfinalcopy 

\begin{document}

\title{Grounding Scene Graphs on Natural Images via Visio-Lingual Message Passing}

\author{Aditay Tripathi$^1$~~~~~~~Anand Mishra$^2$~~~~~~~Anirban Chakraborty$^1$\\
$^1$Indian Institute of Science~~~~~~~~$^2$ Indian Institute of Technology Jodhpur\\
{\tt\small \{aditayt,anirban\}@iisc.ac.in}~~~~~
{\tt\small mishra@iitj.ac.in}\\
\href{https://iiscaditaytripathi.github.io/sgl/}{\textbf{https://iiscaditaytripathi.github.io/sgl/}}
}

\thispagestyle{empty}

\maketitle
\begin{abstract}
   This paper presents a framework for jointly grounding objects that follow certain semantic relationship constraints given in a scene graph. A typical natural scene contains several objects, often exhibiting visual relationships of varied complexities between them. These inter-object relationships provide strong contextual cues towards improving grounding performance compared to a traditional object query-only-based localization task. A scene graph is an efficient and structured way to represent all the objects and their semantic relationships in the image. In an attempt towards bridging these two modalities representing scenes and utilizing contextual information for improving object localization, we rigorously study the problem of grounding scene graphs on natural images. To this end, we propose a novel graph neural network-based approach referred to as \underline{V}isio-\underline{L}ingual \underline{M}essage \underline{PA}ssing \underline{G}raph Neural \textbf{Net}work (\model{}). In \model{}, we first construct a directed graph with object proposals as nodes and an edge between a pair of nodes representing a plausible relation between them. Then a three-step inter-graph and intra-graph message passing is performed to learn the context-dependent representation of the proposals and query objects. These object representations are used to score the proposals to generate object localization. The proposed method significantly outperforms the baselines on four public datasets.
  \end{abstract}

\section{Introduction}
\label{sec:intro}
\epigraph{\small{``What are the mental events that transpire when our eyes alight upon a novel scene? The comprehension that is achieved is not a simple listing of
the creatures and objects. Instead, our mental representation includes a specification of the various relations that exist among these entities."}
}{\textit{--Biederman et al., ~\cite{quote}}}
\begin{figure}[t!]
    \centering
    \includegraphics[width=0.4\textwidth]{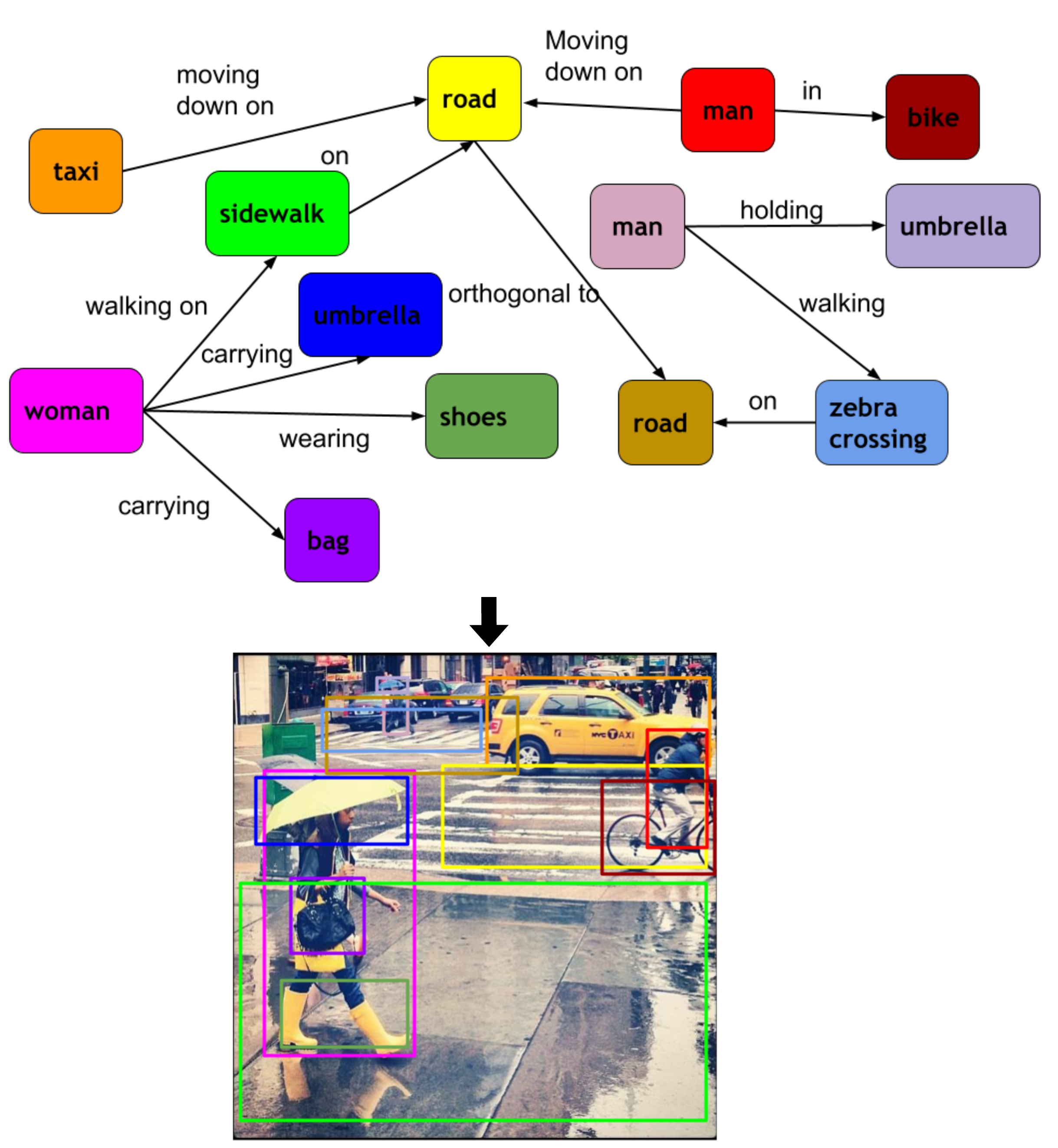}
    \caption{\label{fig:teasure} \textbf{Our goal: Grounding scene graph on image.} Given a scene graph and an image, we ground (or localize) objects and, thereby, indirectly visual relationships as well jointly on the image. \textbf{[Best viewed in color].}}
\end{figure}
The linking of \emph{concepts to context} is referred to as `grounding'~\cite{chandu2021grounding}. In  visual grounding, natural scene is the context, whereas concepts may be expressed using different modalities of queries such as sketch~\cite{Tripathi2020SketchGuidedOL}, natural image~\cite{Hsieh2019OneShotOD}, speech~\cite{verbal}, text~\cite{Kamath2021MDETRM,Li2019VisualBERTAS,Yu2018RethinkingDA} or scene graph~\cite{johnson2015image}. In many computer vision tasks such as image generation~\cite{Lee2020NeuralDN} and image editing~\cite{Su2021FullyFI}, scene graphs have been a popular choice as a query owing to their capability of expressing complex scenes having multiple object instances and semantic relationships among them in a concise, non-ambiguous, and structured way. Further, as noted in~\cite{Ravichandran2021HierarchicalRA}, ``scene graphs explicitly provide a scene’s geometry, topology, and semantics, making them compelling representations for navigation". In fact, scene graphs have proven their utility in embodied AI~\cite{YangWFGM19}, where the scene prior is often encoded as a scene graph, and grounding them in an environment helps embodied agents navigate efficiently.

Foresighting the aforesaid applications, Johnson et al.~\cite{johnson2015image} have introduced the task of scene graph grounding as an auxiliary task to scene graph-based image retrieval as an early work in this direction. While using the graphs as queries for grounding objects is, in principle, exciting, one natural question is how to construct such queries. One possible direction is to use natural language sentences to graph generation~\cite{SchusterKCFM15}, which is yet to achieve an acceptable level of performance for long and complex sentences and is an open area of research~\cite{Wang2018SceneGP}. Another possibility is to use a carefully-designed user interface where non-expert users can quickly draw graphs of arbitrary complexity~\cite{johnson2015image,Zhang2021GeneAnnotatorAS}. Further, one can also obtain scene graph queries by choosing them from a fixed set of scene graphs representing the spatial configuration of objects in a scene obtained using commonsense knowledge~\cite{Giuliari_2022_CVPR} such as \{$\langle$ Monitor, \underline{on}, Table$\rangle$, $\langle$ Keyboard, \underline{near}, Monitor$\rangle$, $\langle$Chair, \underline{next to}, Table$\rangle$, $\langle$Person, \underline{sitting on}, Chair$\rangle$\}. Regardless of the method used to obtain scene graph queries, our scope in this paper is to study scene graph grounding as a standalone task.

In this work, we formulate and study the task of grounding scene graphs on images as defined in Figure~\ref{fig:teasure}, in a principled manner. Further, we propose a novel, robust and effective solution strategy suitable for the query data structure and the task at hand; and provide rigorous experiments and analysis on large-scale computer vision benchmarks. We hope this work will help establish the scene graph grounding as an important and stand-alone cross-modal computer vision problem, thereby leading to exciting contributions toward this open problem.

In this work, we propose a novel method to solve the scene graph grounding problem. To this end, given a query scene graph $G^l$ and an image, we first obtain object proposals on the image using a region proposal network and construct a proposal graph $G^v$. Note that the proposal graph contains object proposals and trainable visual relationship embeddings as representations of nodes and edges, respectively. Now, suppose graph $G^l$ and $G^v$ contain $m$ and $n$ nodes respectively, then we add $m \times n$ auxiliary directed edges between nodes of $G^l$ and $G^v$ to construct a composite visio-lingual graph $G^{vl}$. These auxiliary directed edges allow us to learn the proposal representations relevant to the query graph. We then perform message-passing operations on $G^{vl}$ to learn the contextual proposal and object representation. These learned proposals are scored against each query object to perform visual grounding. We refer to our approach as a \underline{\textbf{V}}isio-\underline{\textbf{L}}ingual \underline{\textbf{M}}essage \underline{\textbf{PA}}ssing \underline{G}raph Neural \underline{\textbf{Net}}work or \model{} in short.

\sloppy
We evaluate~\model{} on four public datasets, namely Visual Genome~\cite{Krishna2016VisualGC}, VRD~\cite{Lu2016VisualRD},  COCO-stuff~\cite{Caesar2018COCOStuffTA}, and SG~\cite{johnson2015image}, and compare it against the following baselines: (i) \emph{Node-only approach} where only the objects in the scene graph query are localized without leveraging the relationship constraints,  (ii) an approach where \emph{flattened triplets obtained from scene graph query} is utilized to perform grounding using a state-of-the-art approach~\cite{Kamath2021MDETRM}. (iii) \emph{CRF-based approach} proposed by~\cite{johnson2015image} where they build a conditional random field (CRF) over the bounding boxes on an image and perform maximum-a-posterior estimation for object localization. These approaches either do not leverage relationships or fail to exploit the structural information present in the graph and thus fall short in performance. Contrary to these approaches, \model{} jointly grounds objects that follow certain semantic relationship constraints given in a scene graph and thereby, outperform them. The implementation for this work is provided at \href{https://iiscaditaytripathi.github.io/sgl/}{https://iiscaditaytripathi.github.io/sgl/}. 

\noindent\textbf{Contributions:} We make the following contributions:
(i) We pose the scene-graph grounding as a standalone problem in a principled manner.
(ii) We propose a novel model -- \model{} towards solving this problem. The \model{} has two novel characteristics. Firstly, a \textit{query-guided proposal graph generation} that utilizes the relationships in the query graph to generate a sparse proposal graph with relevant edges. Secondly, a \textit{visio-lingual message passing network} that learns a query-conditioned structured representation for the object proposals and the query entities to generate better localization.
(iii) We demonstrate efficacy of \model{} via rigorous experiments, ablations, and analysis on modern large-scale public benchmarks.

\section{Related works}
\label{sec:related}
\noindent\textbf{Scene Graph in Computer Vision:} A scene graph is a structured representation of a scene that can precisely and unambiguously represent multiple objects and their semantic relationships. Scene graphs play a pivotal role in holistic scene understanding and are a popular way to represent visual knowledge~\cite{Krishna2016VisualGC}. Being semantically rich in representation, scene graphs have shown their utility in many computer vision tasks such as visual question answering
~\cite{Ghosh2019GeneratingNL,Damodaran2021UnderstandingTR}, image retrieval~\cite{johnson2015image,Wang2020CrossmodalSG}, natural scene generation~\cite{Johnson2018ImageGF,Zhao2019ImageGF,Lee2020NeuralDN}, and high-level image editing~\cite{Dhamo2020SemanticIM,Su2021FullyFI}. In this work, we study scene graphs for grounding multiple objects and relationships jointly on the image.
\begin{figure*}[t!]
\begin{center}
\includegraphics[width=2\columnwidth]{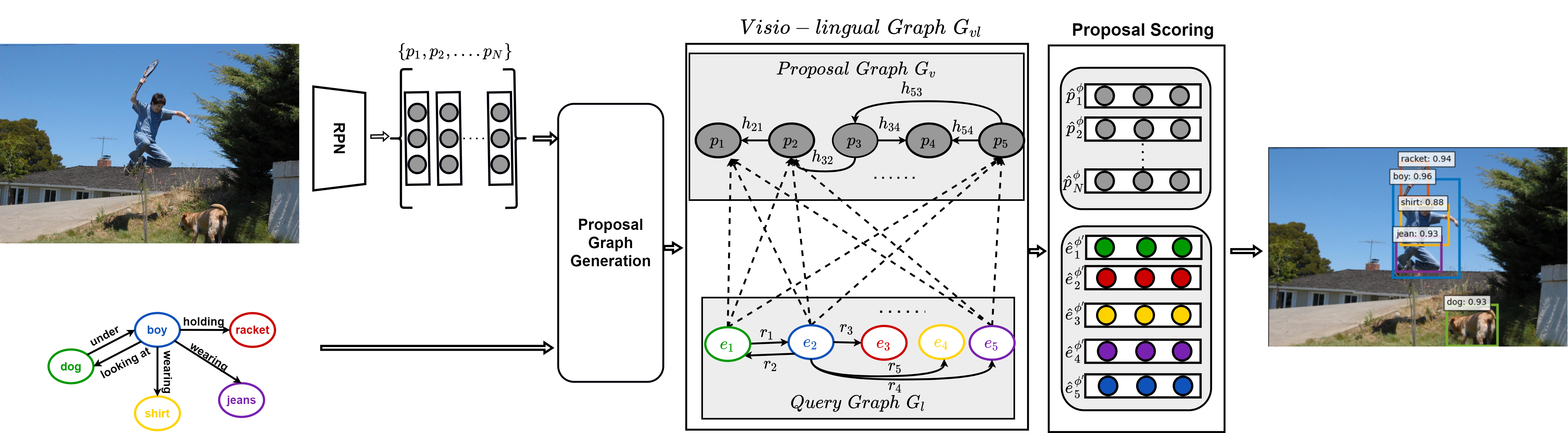}
\end{center}
\caption{\label{fig:overview}
The proposed scene-graph grounding framework (\model{}) works in the following steps: (i) \textbf{Proposal graph generation:} A proposal graph is first constructed using object proposals obtained from RPN as nodes (shown using gray nodes), and edges defined using the relations present in the query. Directed edges from the query nodes to the proposals nodes (shown using dotted arrows) are also included to connect the query and proposal graph (Section~\ref{sec:proposal_graph}). (ii) \textbf{Structured graph learning:} Here, structured representation of proposals and queries are learned by a three-step message passing using edges from the query nodes to the proposals, and in the query and the proposal graphs independently (Section~\ref{sec:srl}), and (iii) \textbf{Proposal scoring} the object proposals are finally scored against the query nodes to localize objects (Section~\ref{sec:proposal_scoring}).
}
\end{figure*}

\noindent\textbf{Query-guided Object Localization:}
In query-guided object localization, the concepts (or queries) that need to be localized on the image are expressed using various modalities in the literature. In~\cite{Tripathi2020SketchGuidedOL} and~\cite{Hsieh2019OneShotOD}, authors use the hand-drawn sketch and natural image of an object, respectively, to express the concept of a `single object'. Cheng et al.~\cite{verbal} use speech input to localize and segment the concept of nouns, i.e., objects and adjectives. Interaction between a subject and an object has been represented by a visual relationship, i.e., $\langle$subject, predicate, object$\rangle$ triplet. The task where both subject and object constrained by a visual relationship, need to be grounded on the image is referred to as Referring Relationship~\cite{Lu2016VisualRD,Krishna2018ReferringR,He2020CPARRCP}. Authors in \cite{Lu2019ViLBERTPT,Li2019VisualBERTAS,Kamath2021MDETRM,Engilberge2018FindingBI,Zhuang2018ParallelAA} use single-line sentence or short phrase as query to ground all the mentioned objects. The idea of scene graphs in computer vision literature has triggered encoding complex semantic concepts (such as the interaction between multiple object categories and instances) in a concise and structured form. Considering this, in a seminal work, Johnson et al.~\cite{johnson2015image} utilized scene graph queries for localizing objects constrained by visual relationships and posed it as a maximum-a-posterior estimation problem. We take this work further by presenting a novel graph neural network-based approach and large-scale evaluation for grounding scene graphs on images.

\noindent\textbf{Graph Neural Networks:} \textcolor{black}{The graph neural network (GNN) was proposed to learn the representation of the entities present in the graph. They have several variants such as graph attention networks~\cite{DBLP:conf/iclr/VelickovicCCRLB18}, graph convolution networks~\cite{DBLP:conf/nips/DefferrardBV16,DBLP:conf/iclr/KipfW17} and message-passing networks~\cite{DBLP:conf/icml/GilmerSRVD17}. Among these, message-passing networks learn the representation for both the nodes and edges in the graph and seen application in many fields such as knowledge graph completion~\cite{DBLP:conf/kdd/0004RL21} , visual relationship detection~\cite{ DBLP:journals/corr/abs-2208-04165}, scene graph generation~\cite{DBLP:conf/eccv/YangLLBP18} and scene understanding~\cite{DBLP:conf/cvpr/Zhang0AT20}. Unlike current literature, we proposed a novel visio-
lingual message passing network to learn the structured representation for the heterogeneous multi-modal graphs.} 
Graph neural network, in general, has been widely used for various scene graph-related tasks. Yang et al.~\cite{Yang2018GraphRF} proposed graph R-CNN for the scene graph generation. They propose an attention-graph convolution network where the global context in the scene graph is used to update object nodes and relationship edge labels. Authors in~\cite{Liu2020LearningCC} use a graph similarity network for grounding \emph{small phrases}. Unlike ours, they assume the availability of natural language query, use its embedding in their framework, and perform message passing independently on visual and lingual graphs. In~\cite{Johnson2018ImageGF}, authors use graph convolution for scene graph-to-image generation; they compute a scene layout by predicting bounding boxes and segmentation masks for objects. Then, they transform the layout into an image with a refinement network. Query-driven proposal graph generation, message passing on multi-modal visio-lingual graphs, and learning context-dependent representation for the proposals and query objects are some highlights of our proposed GNN-based approach, which differentiate us from the scene graph-based GNN literature.


\section{\textbf{V}isio-\textbf{L}ingual \textbf{M}essage \textbf{PA}ssing Graph neural \textbf{NET}work (VL-MPAG Net)}
\label{sec:method}
\subsection{Background and Problem Formulation}
\label{sec:formulation}
A scene graph is a structured representation of the scene containing object instances as its nodes and the relationship between the instances as its edges. Typically, scene graphs also contain object attributes represented as nodes in the graph. However, in the context of this paper, we drop attributes and our scene graphs contain only objects as nodes and relationships as edges. Formally, given a set of object entities $O = \{e_1,e_2,\cdots,e_n\}$ and a set of relations $R = \{r_1,r_2,\cdots,r_m\}$, a scene graph is defined as $s=({\cal V},{\cal E})$ such that ${\cal V} \subset O$ and $\cal{E} \subseteq V\times R'\times V$ is a set of labeled edges, where $R' \subset$$ R$.  

Let $\{I_u, G^l_u\}_{u=1}^{M}$ be the $M$-pairs of natural images and corresponding scene graphs, respectively in a trainset of a dataset. Further, let $O$ and $R$ be the set of all object entities and relations present in the dataset. Each query scene graph $G^l_u$ is a set of triples $\{(e_i,r_j,e_k)\}$ such that the object entities $e_i,e_k \in O$ and the relations $r_j\in R$. During inference, given an image $I_u$ and a scene-graph query $G^l_u$, the task of \textit{grounding scene graphs} on natural images involves localizing objects on the image that correspond to the entities, in the scene graph, that follow the constraints given by the corresponding relations in the scene graph.

The proposed end-to-end trainable framework is illustrated in Figure~\ref{fig:overview}. It works in the following three stages: (i) proposal graph generation, (ii) structured graph learning, and (iii) joint proposal scoring which we describe next. 

\begin{table}[!t]
\scriptsize
\centering
\begin{tabular}{ll}
\toprule
Notation    & Meaning  \\
\midrule
$G^v$, $G^l$, $G^{vl}$ & Proposal graph, query graph, visio-lingual graph 
                           \\
$I_u$, $G^l_u$ & $u^{th}$ natural image and $u^{th}$ scene graph \\
$O$, $R$ & Set of object entities and set of relations \\
${\cal R}_u$ & Set of region proposals for image $I_u$\\
$\Phi_i, i \in\{1,2,\dots,6\}$ & Neural networks\\
$\mathbf{W},\mathbf{W}_i, i\in \{1,\dots,4\}$ & Trainable matrices\\
$e_k,e_j$ & Nodes in query graph $G^l$\\
$r_{kj}$& Edge in query graph between nodes $e_k$ and $e_j$ \\
$p_k,p_j$ & Nodes in proposal graph $G^v$\\
$h_{kj}$& Edge in proposal graph between nodes $p_k$ and $p_j$\\
$S_{kl}$ & Similarity score between proposal $p_k$ and entity $e_l$\\
$ES_{kj,i}$ & Similarity score between edge $h_{kj}$ and relation $r_i$
\\
\bottomrule
\end{tabular}
\caption{Notations used in this paper.}
\end{table}

\subsection{Proposal Graph Generation}
\label{sec:proposal_graph}
Given an image $I_u$, we generate a set of region proposals ${\cal R}_u$ using a region proposal network (RPN) proposed in~\cite{Ren2015FasterRT}. It is a neural network that generates a fixed set of region proposals represented using bounding box coordinates and confidence score of being a foreground, i.e., one of the object categories in the query graph. It should be noted here that RPN does not provide an exact object category label for the generated proposal. One plausible approach for grounding scene graphs could be to score these region proposals against the entity nodes present in the scene graph query to generate object localization. However, this approach does not utilize the structural information in the query graph or the target image and is unlikely to be very effective.

The query scene graph explicitly captures the structural information present among the objects and is represented as the edges that denote the relationships between objects. However, to incorporate the structural information (inter-object semantic relationship) present among different regions in the target image, we create a graph with region proposals as the nodes and the edges as the relationships constraining them. The RPN gives a set of region proposals, and we propose a `query-driven' strategy to establish directed edges between pairs of proposals by leveraging the semantic relations present in the query scene graph. If the proposals are fully connected, the proposal graph would have $O({|{\cal R}_u|^2})$-edges. However, the number of actual connections is restricted by the plausible set of relationships constrained by the visual semantic association between the objects in the scene and is much smaller than $O({|{\cal R}_u|^2})$.

Given a pair of region proposals $(p_k, p_j)$ and their corresponding bounding box coordinates $(B_k, B_j)$, we first estimate the representation of the proposals as follows: $p_k^{\phi} = \Phi_1(p_k)$ and $p_j^{\phi} = \Phi_1(p_j)$,
where $p_k^{\phi}, p_j^{\phi}\in \mathbb{R}^d$ and $\Phi_1$ is a neural network. Note that $p_k$ and $p_j$ denote the image region bounded by $B_k$ and $B_j$ on image respectively. We then compute the representation of the edge between these two nodes as $h_{kj}^{\phi} = \textbf{W}\left[p_k^{\phi},p_j^{\phi},\Phi_2\left(\left[\gamma_{k,j}, \gamma_{k,kj}, \gamma_{j,kj} \right]\right)\right]$
. Here, $\textbf{W}\in \mathbb{R}^{3d\times d}$, $h_{kj}^{\phi}\in \mathbb{R}^d$ and $\Phi_2$ is neural network. The $\gamma$s are computed using bounding boxes $B_k$ and $B_j$ as follows: given a pair of bounding boxes $B_k$ and $B_j$, we first construct a union rectangular box $B_{kj}$ that tightly encloses $B_k$ and $B_j$, and then we compute geometric features for each pair of boxes in $\left\{(B_k,B_j), (B_k, B_{kj}), (B_j, B_{kj})\right\}$. For example, the geometric features for the pair $(B_k, B_j)$ are computed as follows:
\begin{equation}
    \gamma_{k,j} = \left[\ln\frac{|x_k-x_j|}{w_k}, \ln\frac{|y_k-y_j|}{h_k},\ln\frac{w_j}{w_k},\ln\frac{h_j}{h_k}\right]^T,
\end{equation}
where, $(x_k,y_k,w_k,h_k)$ are the bounding box coordinates of the box $B_k$; and 
these features are generated for every pair of object proposals. Now, to retain only those edges for which the representations obtained using visual cues $h_{k,j}^\phi$ are well aligned to at least one of the relations present on the given query scene graph, we score each pair of proposals $(p_k, p_j)$ against the relations present in the scene graph query as follows:
\begin{equation}
\label{eq:cos}
    relSim_{kj} = \max\left(\Theta\left({h_{kj}^{\phi}},r_i\right)\right)_{i=1}^{K},
\end{equation}
where, $\{r_1, r_2, \dots,r_K\}$ is the set of relations present in the query graph $G^l_u$ and $\Theta$ is cosine similarity. Now, we add directed edge from the proposal $p_k$ to $p_j$ if the relationship similarity score $relSim_{kj}$ is above a predefined threshold. This process is repeated for each pair of proposals to generate a directed graph with $p_k^{\phi}$ as the node representations and $h_{kj}^{\phi}$ as the edge representations. Please note that all the mappings in this graph generation process, i.e. $(\textbf{W}, \Phi_1, \Phi_2)$ are learnable and are trained in an end-to-end fashion.

\subsection{Structured Graph Learning}
\label{sec:srl}
Let $G^l_u$ be the directed graph representing the $u^{th}$ scene-graph query, and $G^v_u$ be the corresponding proposal graph generated from image $I_u$ as described in the the previous section (Section~\ref{sec:proposal_graph}). The nodes in the query graph $G_u^l$ represent the objects, and the edges represent the relationship between the object nodes. We use Glove~\cite{Pennington2014GloVeGV} to get the initial representation of the entities and relations in $G_u^l$. The representation of nodes in both of these graphs is updated by passing messages from the neighbors. However, if the proposal representations are updated independently of the query nodes, the same proposal representation will be learned for different query scene graphs. For example, consider an image containing a person wearing a hat and shoes, and two different queries \textit{(person, wearing, hat)} and \textit{(person, wearing, shoes)}. Concretely, in a proposal graph, a region proposal that corresponds to the query node \textit{person} might have neighboring proposals that may correspond to \textit{hat} or \textit{shoes}. If we update the representation of proposal nodes independent of the query nodes, the representation of the proposal gets evenly influenced by all its neighbors, even though some of the neighbors do not correspond to any of the query nodes. To mitigate this problem, we add directed auxiliary edges from each node of the query graph $G_u^l$ to each node of the proposal graph $G_u^v$ as shown by the dotted edges in Figure~\ref{fig:overview} and construct a combined \emph{visio-lingual graph}. A three-step message passing on this graph is performed to update the representation of nodes.

In the first step, message passing is performed on the auxiliary edges from the query graph to the proposal graph, and the representation of object proposals is updated as:
\begin{equation}
    \bar{p}_k^{\phi} = \textbf{W}_1 p_k^{\phi} + \textbf{W}_2\sum_j sim_{kj}\cdot e_{j}^{\phi'},
\end{equation}
\begin{equation}
    sim_{kj} = \frac{e^{\left[\left(\textbf{W}_3 p_k^{\phi}\right)^T\left(\textbf{W}_4 e_j^{\phi'}\right)\right]}}{\sum_l e^{\left[\left(\textbf{W}_3 p_k^{\phi}\right)^T\left(\textbf{W}_4 e_l^{\phi'}\right)\right]}},
\end{equation}
where, $\textbf{W}_1, \textbf{W}_2, \textbf{W}_3, \textbf{W}_4 \in \mathbb{R}^{d\times d}$ and $e_j^{\phi'}$ is the representation of entity $e_j$ in the query graph obtained using Glove. In this step, a weighted sum of the representation of the query nodes is added to each proposal's representation. Also, the weight depends on the compatibility of the proposal representation with the query nodes. This step helps incorporate query information in the proposal representation. In other words, for a target image, the final representation of the region proposals will be learned differently for each query. 

In the second step, message passing is performed on the query scene graph. For the query graph $G_u^l$, the node representations are updated as follows:
\begin{equation}
    \widehat{r}_{kj}^{\phi'} = \Phi_3\left(\left[e_k^{\phi'},e_j^{\phi'},r_{kj}^{\phi'}\right]\right),
\end{equation}
\begin{equation}
    \widehat{e}_k^{\phi'} = \frac{1}{|nbd({e_k})|}\sum_{nbd({e_k})} \Phi_4\left(\left[e_k^{\phi'},\widehat{r}_{kj}^{\phi'}\right]\right),
\end{equation}
where $nbd({e_k})$ is the set of neighbouring nodes of $e_k$ in the graph $G_u^l$ and $\Phi_3, \Phi_4$ are two-layer neural networks. Further, in the third step, given the proposal graph $G_u^v$, the representation of the proposal nodes are updated as follows:
\begin{equation}
    \widehat{h}_{kj}^{\phi} = \Phi_5\left(\left[\bar{p}^{\phi}_k, \bar{p}^{\phi}_j, h_{kj}^{\phi}\right]\right),
\end{equation}
\begin{equation}
    \widehat{p}_k^{\phi} = \frac{1}{|nbd({p_k})|}\sum_{nbd({p_k})} \Phi_6\left(\left[\bar p_k^{\phi},\widehat{h}_{kj}^{\phi}\right]\right).
\end{equation}
Here, $nbd({p_k})$ is the set of neighbours of node representing proposal $p_k$ in the proposal graph $G_u^v$, and $\Phi_5, \Phi_6$ are two-layer neural networks. After learning the contextual representation of nodes in both graphs, the proposal nodes are scored against the query nodes to ground the scene graph on the image. The representations learned after the second and third steps of message passing can be made more expressive by using two layers of GNN because it enables the model to utilize a 2-hop neighborhood context.

\subsection{Joint Proposal Scoring}
\label{sec:proposal_scoring}
Once the representation of the query objects and the region proposals are updated, a scoring function $\Theta$ is used to score the region proposals with the query objects. Consider a proposal $p_k\in {\cal R}_u$ with a label variable $y_k$. During the training phase, $y_k$ is assigned a class $c_{e_l}$ or $0$ based on its intersection-over-union (IoU) with the ground truth bounding box of the query object $e_l$ belonging to class $c_{e_l}$. It is assigned the class $c_{e_l}$ when its IoU $\geq$ $0.5$ and $0$ otherwise.
Once the labels are assigned to the proposal boxes, the score of each region proposal with respect to the queries are generated as follows: $S_{kl} = \Theta(\widehat{p}_k^{\phi}, \widehat{e}_l^{\phi'})$, where $\Theta$ is cosine similarity and $S_{kl}$ is the similarity score between representations of the proposal $p_k$ and query node $e_l$. For each node $e_l$ in the query graph and a set of region proposal ${\cal R}_u$ for image $I_u$, the loss function is defined as follows:
\begin{equation}
\begin{split}
    L(Q_u, e_l) =& \sum_k\Big\{-\left(\mathbbm{
    1}_{[y_k=c_{e_l}]} 
    \ln(S_{kl})\right)\\&-
    \left(\mathbbm{
    1}_{[y_k\neq c_{e_l}]} \ln(1-S_{kl})\right) + L_{MR}^k\Big\}.
\end{split}
\end{equation}
Here, $L^k_{MR}$ is a margin loss and is defined as follows:
\begin{equation}
\label{eq:margin}
    \begin{split}
    L_{MR}^k =& \sum_{j=k+1} \Big\{\mathbbm{1}_{[y_{k}=y_{j}]}max(|S_{kl}-S_{jl}|-m^-\\
            &,0) + \mathbbm{1}_{[y_{k}\neq y_{j}]}max(m^+ -|S_{kl}-S_{jl}|,0)\Big\},
    \end{split}
\end{equation}

where $m^+$ and $m^-$ are the positive and negative margins, respectively, and $y_k$ is the class label for the proposal $p_k$. The margin loss in Equation~(\ref{eq:margin}) takes a pair of proposals and ensures that the proposal pair that are assigned the same label has prediction probabilities closer to each other and at the same time makes the proposals with different labels wider in terms of prediction probabilities.

\begin{table}[!t]
\scriptsize
\renewcommand*{\arraystretch}{1}
\centering
\begin{tabular}{lcc|cc|cc}
\toprule
{Model}              & \multicolumn{2}{c|}{COCO-stuff} & \multicolumn{2}{c|}{VG-FO} & \multicolumn{2}{c}{SG} \\
                                    &R@1            &R@5           &R@1         &R@5&R@1        &R@5\\
\midrule
\textbf{Edges removed} &          &          &         &         &        &       \\
~~~\textit{node-only (Detection)}                  & 21.0           & 47.9          & 30.1        & 62.8        & 23.4       & -         \\
~~~\textit{node-only (Localization)}                   & 33.9           & 57.2          & 29.9        & 53.5        & 34.7       & 62.5      \\
\textbf{Flattened triplets} &          &          &         &         &        &       \\
~~~MDETR~\cite{Kamath2021MDETRM}                               & 30.1           & 47.9          & 25.4        & 44.8        & $15.9$       & $29.9$      \\
\textbf{Structured Graph Query} &          &          &         &         &        &       \\
~~~CRF-Based$^*$~\cite{johnson2015image}                                & -              & -             & -           & -           & 23.9       & -         \\
~~~\textbf{Ours (\model{})} &          &          &         &         &        &       \\
~~~~1-layer  & 35.5           & 57.9          & 32.7        & 61.6        & 35.9       & 64.2      \\
~~~~2-layers & \textbf{36.3}           & \textbf{58.4}          & \textbf{36.0}        & \textbf{63.3}        & \textbf{36.9}       & \textbf{65.6}      \\
\bottomrule
\end{tabular}
\caption{\label{tab:full}\textbf{Results for scene graph grounding task on COCO-stuff val and VG-FO for completely overlapping train-test categories setting.} $^*$Due to the unavailability of the implementation of~\cite{johnson2015image} at the time of submission of this paper, we only compare with reported results in their paper.}
\end{table}
To select a desirable set of edges during proposal graph generation, a loss function is also defined on edges that can be constructed from the set of region proposals. For a set of $N$ region proposals, $N \choose 2$ edges connecting a pair of region proposals can be defined.  Let ${\cal E}$ be the set of all such edges. Consider a directed edge $h_{kj} \in {\cal E}$ (where $k$ and $j$ are source and target nodes respectively.) and its label variable $z_{kj}$. The variable $z_{kj}$ is assigned a label $c_{r_i}$ if the pair of proposals corresponding to the edge $h_{kj}$ follows the relation $r_i$ in the query graph. The visual representation of the said edge is subsequently scored against the relationship embedding $r_i$ as follows: $ES_{kj,i} = \Theta({h}^{\phi}_{kj},{r}^{\phi'}_i)$,
where, $\Theta$ is cosine similarity, and $ES_{kj,i}$ is the score between the edge $h_{kj}$ and the relation $r_i$. 
For the relation $r_i$ and the set of edges ${\cal E}$, the loss is defined as follows:
\begin{equation}
\label{eq:rel_loss}
\begin{split}
    L({\cal E}, r_i) =& \sum_l\Big\{-\left(\mathbbm{
    1}_{[z_{kj}=c_{r_i}]} \ln(ES_{kj,i})\right)\\ &-
    \left(\mathbbm{
    1}_{[z_{kj}\neq c_{r_i}]} \ln(1-ES_{kj,i})\right) \Big\},
\end{split}
\end{equation}

where, $c_{r_i}$ is the label of relation $r_i$. One example of such a label is `\emph{wearing}'. Generally, for a relation, the number of positive and negative edges have a huge imbalance (usually much more negative than positive edges), leading to poor training. To mitigate this problem, we present the following strategy to sample a more balanced set of edges.

Consider an edge $(l,m)$ (relation) in a query scene graph. All $N$ region proposals are scored, as defined previously, with both the nodes ($e_l$ and $e_m$) present in the edge. 
Suppose $P_l$ and $P_m$ are the list of proposal sorted in decreasing order of scores with respect to $e_l$ and $e_m$. 
We select $p$ proposals each from $P_l$ and $P_m$ randomly but ensure half of them come from the top 50 of each list. From these sets of $p$ proposals, a set of edges are formed that connect proposals from selected lists. These steps are repeated for all the edges in the query scene graph to obtain a balanced subset. Then, the loss function defined in Equation~(\ref{eq:rel_loss}) is computed for these balanced subsets of edges in mini-batches. In our experiments, we empirically choose $p=48$. We also define cross-entropy loss on
the labeled (foreground or background) feature vectors of the region proposals and a regression loss on the predicted bounding box locations with respect to the ground truth bounding boxes.

\noindent\textbf{Inference:} During inference, after obtaining object proposals on the image, we update query objects and proposals embeddings via message passing on the constructed graph. Different from the training, we then score the region proposals against the query objects and choose the highest-scoring proposals for each query object as the localization output.

\section{Experiments and Results}
\subsection{Datasets, Evaluation Protocols and Baselines}
\label{sec:dataset}
We use four public datasets, namely 
Visual Genome~\cite{Krishna2016VisualGC}, VRD~\cite{Lu2016VisualRD}, 
COCO-stuff~\cite{Caesar2018COCOStuffTA} and SG~\cite{johnson2015image} for our experiments. Among these, motivated by VRR-vg~\cite{Liang2019VrRVGRV} and VG-150~\cite{Xu2017SceneGG}, to minimize the bias due to long-tail distribution and visually-irrelevant relationship (such as a \emph{field for plane} or \emph{sign that says pumpkin}), we use a subset of the visual genome containing 93K image-scene graph pair for training and 40K image-scene graph pair for testing. The scene graphs in this dataset are constructed using 150 object categories and 40 predicates. We refer to this split of the visual genome as \emph{Visual Genome-Fully Observed (or VG-FO)}. To facilitate the study of grounding unseen objects, we create a split called \emph{Visual Genome-Partially Observed (or VG-PO)} which contains scene graphs constructed from subsets of 125 object categories during training, whereas the testing scene graphs contain subsets of additional 25 object categories. The other three datasets, i.e., VRD~\cite{Lu2016VisualRD}, COCO-stuff~\cite{Caesar2018COCOStuffTA}, and SG~\cite{johnson2015image} contain (45K,~100,~70), (77K,~183,~6), and (5K,~166,~68) number of images-scene graph pairs, object categories, predicates in all. The scene graphs for COCO-stuff are constructed using protocols from~\cite{Johnson2018ImageGF}. We use Recall at 1 and 5 (denoted as R@1 and R@5 from here onwards) to evaluate scene graph grounding by considering an object localization as correct when its intersection over union with the ground truth bounding box is $\geq$ $0.5$
\begin{table}[!t]
\tiny
\renewcommand*{\arraystretch}{1.0}
\centering
\begin{tabular}{l|l|rrrrrrrr}
\toprule
\multicolumn{2}{l}{Edges}     &~~1~~&~~2~~&~~3~~&~~4~~&~~5~~&~~6~~&~~7~~&~~8~~    \\
\midrule
\multirow{3}{*}{VG-FO}        & R@1       & 33.8  & 35.8 & 33.5 & 30.7 & 27.9 & 26.0 & 24.6 & 23.2 \\
                              & R@5       & 62.7  & 64.9 & 61.2 & 58.2 & 54.2 & 53.2 & 49.5 & 48.6 \\
                              & \#Samples & 21,807 & 7,543 & 3,826 & 2,347 & 1,547 & 936  & 693  & 434  \\
\hline
\multirow{3}{*}{VG-PO-Unseen} & R@1       & 23.9  & 30.0 & 33.7 & 35.1 & 34.7 & 35.7 & 40.1 & 35.7 \\
                              & R@5       & 51.2  & 58.1 & 61.5 & 59.8 & 56.1 & 56.4 & 59.2 & 56.9 \\
                              & \#Samples & 25,913 & 7,665 & 3,005 & 1,391 & 770  & 390  & 218  & 130 \\
\bottomrule
\end{tabular}
\caption{\label{tab:edges} \textbf{Effect of the size of the query on the performance of the model.} Unseen refers to the set of categories in VG-PO not used during training. (Refer to Section~\ref{sec:results}).}
\end{table}
\begin{figure*}[!t]
\begin{center}
\includegraphics[width=\linewidth]{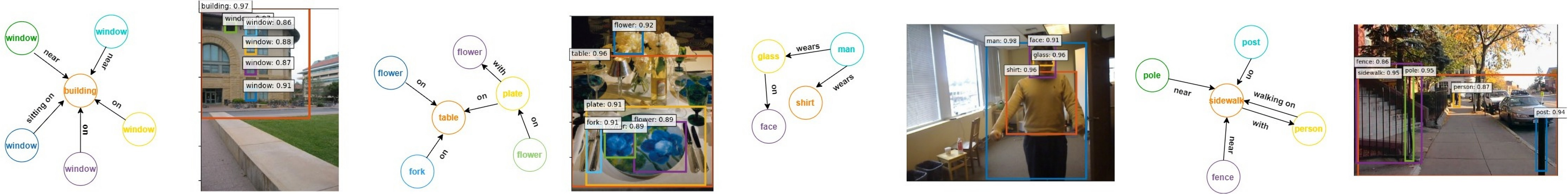}
\end{center}
\caption{\label{fig:visuals} A selection of results on Visual Genome. Scene graph query along with the grounding results are shown side by side using the same color border for object nodes in the query graph and corresponding grounded object bounding boxes. 
}
\end{figure*}

\noindent\textbf{Baselines:}
\label{sec:baselines}
As there is no existing method that addresses the task of visual grounding when scene graph is used as query except a CRF-based approach~\cite{johnson2015image}. Therefore, along with comparing against them, we present baselines to understand: (i) \textbf{Importance of visual relations (i.e., edges in the query) in localizing objects.} To this end, we present the following two baselines for edges-removed queries: 
(a) \emph{Node only (Detection-based):} 
We detect the set of object categories in the query scene graph using Faster-RCNN~\cite{Ren2015FasterRT}.
Note that this model is limited to object categories seen during training. (b) \emph{Node only (Localization-based):} In this, we obtain region proposals using faster-RCNN and then score them against the Glove word representation of each object present in the query graph to generate localizations. (ii) \textbf{Importance of structured property of the query graph.} To this end, we use flattened triplets (subject-predicate-object) obtained from the scene graph as a query in MDETR~\cite{Kamath2021MDETRM}
-- a transformer-based model for visual grounding task. To make a fair comparison with our model, we train this model only on our datasets without any pretraining and use Resnet50 as the backbone. Further, when the scene graph contains only two nodes, the problem of scene graph grounding reduces to referring relationships~\cite{Krishna2018ReferringR}. Thus, for such cases, we compare with state-of-the-art referring relationship methods~\cite{Krishna2018ReferringR,Lu2016VisualRD,He2020CPARRCP}.


\subsection{Results and Discussion}
\label{sec:results}
We first show the results on VG-FO, COCO-stuff, and SG datasets in Table~\ref{tab:full}. We observe that~\model{} outperforms all the baselines on all datasets. 
The node-only baselines do not leverage the visual relationship in the scene graph query and perform poorly. The flattened scene-graph grounding approach (MDETR) does not encode the structural information in a scene graph and falls short in performance. Also, it needs to deal with language understanding challenges such as co-references, noun phrase and relationship extraction, and long-range dependency of the concepts. MDETR requires a lot of training data; therefore, to evaluate the model on the SG dataset (which contains only 4K training samples), we utilize the MDETR model trained on the VG-FO dataset. The~\model{} outperforms the CRF-based approach~\cite{johnson2015image}, indicating better representation learning by GNNs than CRF for the scene-graph localization task. Further, COCO-stuff has fine-grained object categories that are semantically very close (e.g., \emph{wall-wood} vs. \emph{wall-stone}). This causes inferior performance of \emph{node only (detection)} baselines on COCO-stuff.
\begin{table}[!t]
\scriptsize
\renewcommand*{\arraystretch}{1}
\centering
\begin{tabular}{l|cc|cc}
\toprule
\multirow{2}{*}{Model} & \multicolumn{2}{c|}{Subject} & \multicolumn{2}{c}{Object} \\
                       &~~~~R@1~~~~&~~~~R@5~~~~&~~~~R@1~~~~&~~~~R@5~~~~               \\
\midrule
SSAS~\cite{Krishna2018ReferringR}              & 21.5      & -           & 24.2       & -           \\
VRD-LP~\cite{Lu2016VisualRD}               & 31.5      & 38.8         & 34.9       & 40.3           \\
CPARR~\cite{He2020CPARRCP}             & 49.8      & 69.4          & \textbf{52.4}       & 70.2        \\
\textbf{Ours (\model{}})         & \textbf{51.6}      & \textbf{79.3}          & 51.7      & \textbf{76.1}          \\
\bottomrule
\end{tabular}
\caption{\label{tab:vrd} \textbf{Comparison of \model{} against the referring relations baselines} for the scenario when graph contains only two nodes on VRD dataset.}
\end{table}

\noindent\textbf{Effect of number of edges in the query graph:}
We perform grounding scene graphs experiments with varying sizes of the query scene graph. The largest scene graphs we ground on the image in our experiments contain eight edges. To analyze the performance of \model{} with respect to scene graph size, we compute R@1 and R@5 with varying numbers of edges in the query scene graph on two splits of VG datasets used in this paper. As shown in Table~\ref{tab:edges}, our method successfully grounds scene graphs even when the graph contains as large as eight edges. In the case of VG-FO, as the dataset contains fewer samples of large-size scene graphs, there is a drop in recall when the graph size becomes larger. In contrast, on grounding unseen objects (also refer to Grounding Unseen Objects towards the end of this section), a larger scene graph size helps. This result is intuitive as a large graph gives better global context, subsequently enabling the grounding of unseen objects as well.

For scene graphs containing only one edge, the problem of scene graph grounding reduces to referring relationship. We directly compare our approach with state-of-the-art referring relationships methods in Table~\ref{tab:vrd} on VRD dataset. Here, CPARR~\cite{He2020CPARRCP} utilizes the relation between the query nodes by combining the node prediction score with the relation prediction score at the last stage. \model{}, instead, utilizes the relation information during the initial stages of modeling and, thereby, achieves competitive if not better R@$1$ and significantly better R@$5$ (nearly 10\% and 6\% better compared to the most competitive method).

\noindent\textbf{Qualitative Analysis:} A selection of scene graphs grounding on the VG-FO dataset is shown in Figure~\ref{fig:visuals}. From detailed analysis (refer to the supplementary material), we observe that \model{} is able to localize the correct objects in a dense image containing instances of many object categories. As an example in the Figure~\ref{fig:visuals}, in the second example, the model is able to localize the `flowers' printed on the `plate' as specified in the query. 

\noindent\textbf{Multi-instance Localization:}
The representations learned for the nodes of the same class in our framework differ due to the difference in the one- or two-hop neighboring objects and relations. Therefore, our framework naturally enables multi-instance localization. Even if one- or two-hop objects and relations are the same (For example, the leftmost localization in Figure~\ref{fig:visuals}), our model allows localizing all the objects corresponding to each node and thereby enables multi-instance localization. However, in such rare cases, it becomes infeasible to disambiguate different instances. 

\noindent\textbf{Grounding Unseen Objects:}
\label{subsec:unseen}
Although the primary goal of this work is to address the task of grounding scene graphs on natural images, localization of unseen object categories is an auxiliary but
challenging setting that we also evaluated on the VG-PO dataset in Table~\ref{tab:unseen_vg}. Compared to `seen', `unseen' object categories suffer in performance. However, the proposed model can better capture the context between objects, leading to significantly better performance than the baselines. The \emph{node only (loc.)} suffers the most because it does not utilize the relations that might help the localization of `unseen' object categories. MDETR, on the other hand, uses a transformer-based model to learn the contextual embeddings and captures the context for `unseen'  categories, which in turn aid in their localization. \textit{Node only (det.)} requires all the object categories to be known during training; hence, is dropped for comparison.
\begin{table}[!t]
\scriptsize
\centering
\begin{tabular}{lcc|cc}
\toprule
\multirow{2}{*}{Model} & \multicolumn{2}{c|}{~~~~~~Seen Categories~~~~~~} & \multicolumn{2}{c}{~~~~~~Unseen Categories~~~~~~} \\
                       &~~~~R@1~~~~&~~~~R@5~~~~              &~~~~R@1~~~~&~~~~R@5~~~~       \\
\midrule
\textit{Node only (Localization)}              & 33.2      & 56.6           & 19.6       & 43.1            \\
MDETR~\cite{Kamath2021MDETRM}                  & 26.2      & 47.1           & 26.4       & 45.7            \\
\textbf{Ours(\model{})}    &&&&\\
~~~~~1-layer                  & 38.0      & 64.9      & 27.5       & 54.5      \\
~~~~~2-layers  & \textbf{39.9} & \textbf{66.9}  & \textbf{29.0} & \textbf{53.6}\\
\bottomrule
\end{tabular}
\caption{\label{tab:unseen_vg}  The proposed model outperforms the baselines for `seen' and `unseen' object categories on VG-PO dataset.}
\end{table}

\noindent\textbf{Robustness to Sparse and Incomplete Query:}
The scene graph grounding method must be robust against sparse and incomplete queries, not just clean ones. To demonstrate the robustness of the proposed model, we first perturb the scene-graph queries by introducing different degrees of noise and then evaluate our model against such perturbed queries. For each edge in the scene graph, we perturb the graph with a probability $p$ by replacing the subject, object, or relation with synonyms obtained using the WordNet synsets or removing the relation from the query graph. We utilized the \model{} model trained on the VG-FO dataset to perform this analysis. For 10\% to 40\% noise, we obtain the $R@1=[30.6, 29.9, 29.0, 28.2]$ respectively, demonstrating the robustness of our model to incomplete and sparse scene graph queries.



\begin{table}[]
\scriptsize
\centering
\begin{tabular}{ccc|ccc}
\toprule
AE-MP & QG-MP & PG-MP & Order & R@1  & R@5  \\
\midrule
      &       &       &              & 29.9 & 53.5 \\
      & \checkmark & \checkmark &              & 28.5 & 53.0 \\
\checkmark &       & \checkmark &              & 31.8 & 59.6 \\
\checkmark &         &       &              & 29.9 & 53.7     \\
\checkmark & \checkmark & \checkmark & QG-MP $\rightarrow$  AE-MP $\rightarrow$ PG-MP        & 31.3 & 59.3 \\
\midrule
\checkmark & \checkmark & \checkmark &        AE-MP $\rightarrow$  QG-MP $\rightarrow$ PG-MP  & \textbf{32.7} & \textbf{61.6} \\

\bottomrule
\end{tabular}
\caption{\label{tab:model_comps} \textbf{Analysis of the message passing steps performed on the VG-FO dataset.} Here AE-MP, QG-MP, and PG-MP denote Message Passing on auxiliary visio-lingual edges, the query graph, and the proposal graph, respectively. $A \rightarrow B$ indicates that $A$ is performed before $B$. The last row represents our full model.}
\end{table}

\noindent\textbf{Analysis of the message-passing steps:}
The message passing, in our framework, is performed on the auxiliary edges from the nodes of the query graph to the nodes of the proposal graph (AE-MP), the query graph (QG-MP), and the proposal graph (PG-MP). To better understand the effect of these message-passing steps and their order, we performed an ablation experiment by removing them one by one and changing the order of AE-MP and QG-MP on the VG-FO dataset. Results of this ablation are reported in Table~\ref{tab:model_comps}. 
The first step, where message passing is performed on the auxiliary edges from the query graph to the proposal graph (AE-MP), is essential, as evident by the decrease in performance (row 2 in Table~\ref{tab:model_comps}) when it is excluded. Further, the message passing on the query graph (QG-MP) and the proposal graph (PG-MP) helps to incorporate the structural information while learning the representation of the nodes and the edges, leading to improvement in localization performance. The order of message passing is also crucial. Performing message passing on auxiliary nodes first, followed
by query graph and proposal graph respectively (AE-MP $\rightarrow$  QG-MP $\rightarrow$ PG-MP) helps \model{} learn better conditional node representations for the proposal graph, thereby helping to achieve better localization. This is also evident in results where we observe that the above order gives a superior performance compared to one where message passing on query graph (QG-MP) is performed before message passing on auxiliary nodes (AE-MP).
\section{Conclusions}
We thoroughly studied the problem of grounding scene graphs on natural images, presented an end-to-end visio-lingual message passing-based graph neural network framework, and performed experiments on large-scale image datasets. The performance improvement over baselines confirms the efficacy of the proposed \model{} and exhibits that better modeling of the context present in scene graphs leads to better grounding. We believe this work will revive research interests and future contributions toward the under-explored scene graph-based grounding problem. 
\newline
\newline
\noindent\textbf{Acknowledgments:}
This work is partially supported by Startup Research Grant (SRG) by the SERB, Govt. of India (File number: SRG/2021/001948) to Anand Mishra.
{\small
\bibliographystyle{ieee_fullname}
\bibliography{egbib}
}
\newpage
\appendix
\section{Dataset details}
\begin{table}[h!]
\small
\centering
\begin{tabular}{lccc}
\toprule
Dataset&\#Samples&\#Predicates&\#Categories\\
\midrule
VG-FO      & 93,323/40,124         & 40/40           & 150/150                  \\
VG-PO      & 69,917/39,699         & 40/40           & 125/150                  \\
VRD        & 39,699/6,869          & 70/70           & 100/100                  \\
COCO-stuff & 74,121/3,074          & 6/6            & 183/183 \\
SG         & 4,000/1,000           & 68/68          & 166/166 \\
\bottomrule
\end{tabular}
\caption{Train/test statistics of the dataset used in our evaluation.}
\end{table}
\section{Scalability Analysis}
Given $N$ entities on the query graph and a set of object proposals ${\cal R}_u$ on the image containing $M$ proposals, the augmented directed graph between proposal and query graph takes $O(MN)$-space. However, this complexity is nearly-linear with respect to the number of proposals ($M$) since even for complex queries $N \ll M$. Further, in theory, if the region proposals are fully connected, the proposal graph would take $O(M^2)$-space. However, in reality, the number of actual connections is restricted by the plausible set of relationships constrained by the visual semantic association between the objects in the scene. In fact, in our experiments, the actual number of edges in the proposal graph is just a tiny fraction of $M^2$ (refer Table~\ref{tab:edges_supp}). 

Further, to illustrate the scalability of our proposed model with respect to the number of edges in the query graph, we compute the average inference time (in seconds) on NVIDIA Quadro-8000. The inference time does not increases significantly as the number of edges in the query graph increases from 1 to 8 (Refer Table~\ref{tab:inf_time}), with a significant under-use of the GPU memory.

\begin{table}[!t]
\centering
\begin{tabular}{llr}
\toprule
Proposals(${\cal R}_u$)&Average number of edges&$|{\cal R}_u|^2$\\
\midrule
128 & 674.2&16,384 \\
256 & 1,444.6 & 65,536 \\
512 & 2,784.4 & 262,144 \\
1024&5,051.9& 1,048,576\\
\bottomrule
\end{tabular}
\caption{\label{tab:edges_supp} The number of edges in the proposal graph is averaged over all the queries in the test set of VG-FO dataset. The average number of edges in the proposal graph selected by \model{} is significantly smaller than $|{\cal R}_u|^2$.}
\end{table}

\begin{table}[]
\small
\centering
\begin{tabular}{lcccccccc}
\toprule
\#Edges&1&2&3&4&5&6&7&8\\
\midrule
Time (s) & 0.12 & 0.12 & 0.12 & 0.12 & 0.13 & 0.13 & 0.13 & 0.14 \\   
\bottomrule
\end{tabular}
\caption{\label{tab:inf_time} The Average inference time (shown in seconds) increases very slowly with the number of edges in the query graph. It shows that the proposed model is scalable with respect to size of the query graph.}
\end{table}
\section{Limitations}
 Although \model{} does not impose any theoretical limitations on grounding very-large-size scene graphs; our work is limited by availability of a balanced dataset with dense scene graph annotations. This limits us to show results of grounding scene graph with maximum of eight-edges only.
 \section{Implementation Details}
 We implemented the model using the PyTorch v1.9.1 and PyTorch-geometric libraries with CUDA $10.2$. The stochastic gradient descent with a momentum of $0.9$ is used to train the models on one NVIDIA-Quadro RTX-8000. The model is end-to-end trained with the learning rate as $0.0001$ and $0.02$ for the fasterRCNN backbone and the remaining network, respectively with a decay by a factor of $0.1$ after every $2$ epoch with a maximum of $10$ epochs. In our experiments, we use $m^+=0.3$ and $m^-=0.7$ in Equation 10. The implementation of this work is available at \href{https://iiscaditaytripathi.github.io/sgl/}{\textbf{https://iiscaditaytripathi.github.io/sgl/}}. We also provided splits for two representative datasets used in this paper.

\section{Additional Visual Results}
Please refer to the following two pages for additional visual results of the proposed model. 
\begin{figure*}[h!]
\begin{center}
\includegraphics[width=2.0\columnwidth]{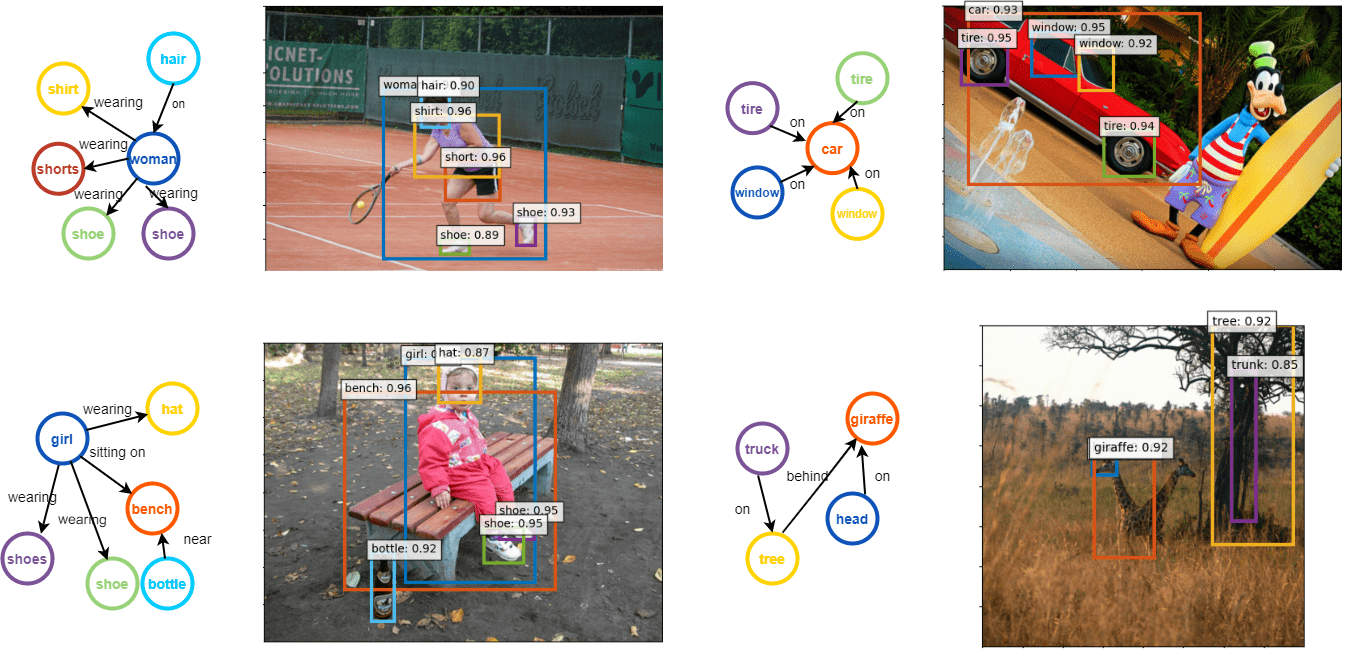}
\caption{More results: Grounding Scene Graphs on Image.[\textbf{Best viewed in color}]}
\end{center}
\end{figure*}

\clearpage
\begin{figure*}[h]
\begin{center}
\includegraphics[width=2.0\columnwidth]{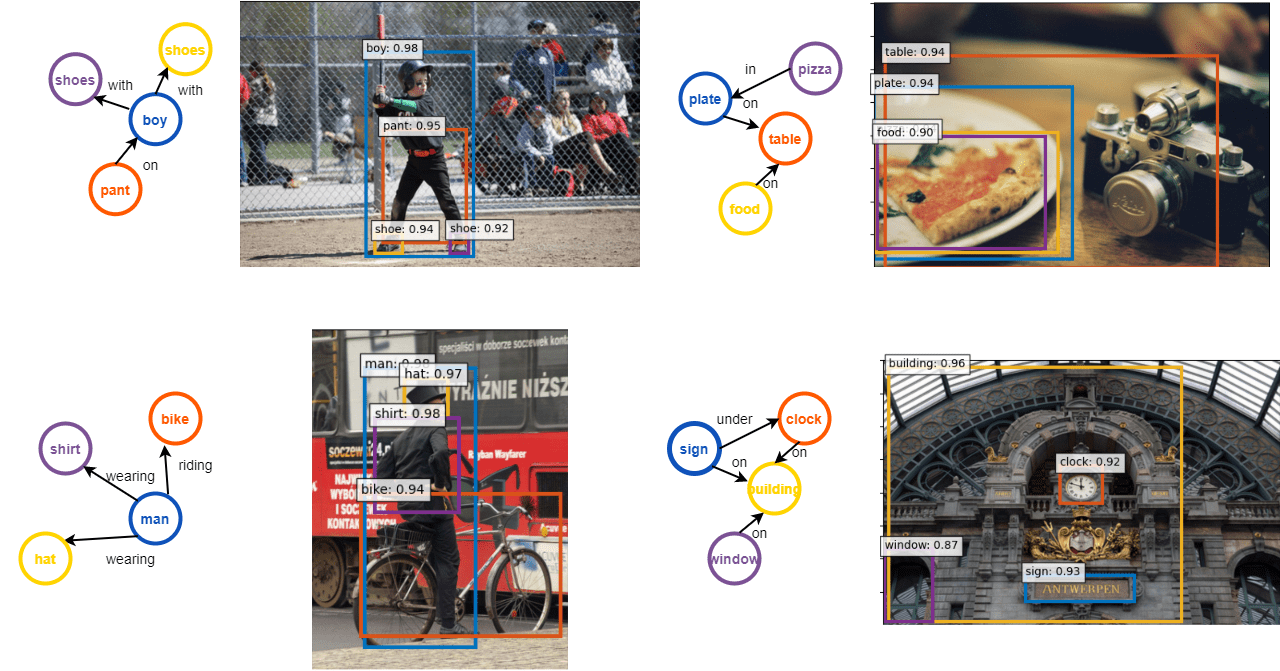}
\end{center}
\caption{More results: Grounding Scene Graphs on Image.[\textbf{Best viewed in color}]}
\end{figure*}

\begin{figure*}[h]
\begin{center}
\includegraphics[width=2.0\columnwidth]{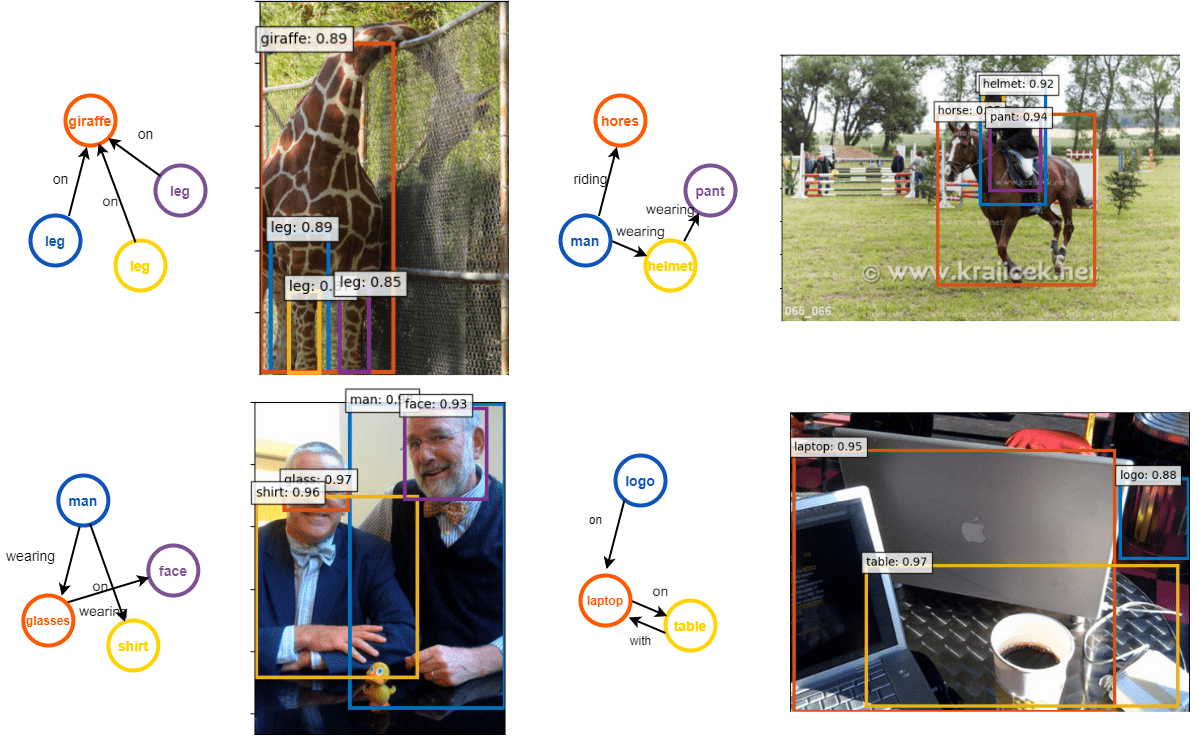}
\end{center}
\caption{More results: Grounding Scene Graphs on Image. Last row correspond to some of the failure cases. In the first the model is confusing between two men and in the second image, the model is not able to detect the logo.[\textbf{Best viewed in color}]}
\end{figure*}

\end{document}